\documentclass[10pt,twocolumn,letterpaper]{article}

\usepackage{cvpr}
\usepackage{times}
\usepackage{epsfig}
\usepackage{graphicx}
\usepackage{amsmath}
\usepackage{amssymb}
\usepackage{gensymb}
\usepackage[pagebackref=true,breaklinks=true,letterpaper=true,colorlinks,bookmarks=false]{hyperref}

\usepackage{authblk}
\usepackage{caption}
\usepackage{subcaption}
\usepackage{wrapfig}

\cvprfinalcopy % *** Uncomment this line for the final submission

\ifcvprfinal\fi

\graphicspath{ {images/} }

\begin{document}

%%%%%%%%% TITLE
\title{Training Deep Neural Networks to Detect Repeatable 2D Features Using Large Amounts of 3D World Capture Data}
\author{Alexander Mai \\
University of California, San Diego \\
9500 Gilman Dr, La Jolla, CA 92093 \\
{\tt\small atm008@ucsd.edu}
\and
Joseph Menke \\
University of California, Berkeley \\
Berkeley, CA \\
{\tt\small joemenke@berkeley.edu}
\and
Allen Yang \\
University of California, Berkeley \\
Berkeley, CA \\
{\tt\small yang@eecs.berkeley.edu}
}
%This work was supported in part by NSF awards CNS-1730158, ACI-1540112, ACI-1541349, OAC-1826967, the University of California Office of the President, and the University of California San Diego's California Institute for Telecommunications and Information Technology/Qualcomm Institute. Thanks to CENIC for the 100Gpbs networks
\maketitle

\begin{abstract}
Image space feature detection is the act of selecting points or parts of an image that are easy to distinguish from the surrounding image region. By combining a repeatable point detection with a descriptor, parts of an image can be matched with one another, which is useful in applications like estimating pose from camera input or rectifying images. 
Recently, precise indoor tracking has started to become important for Augmented and Virtual reality as it is necessary to allow positioning of a headset in 3D space without the need for external tracking devices. Several modern feature detectors use homographies to simulate different viewpoints, not only to train feature detection and description, but test them as well. The problem is that, often, views of indoor spaces contain high depth disparity. This makes the approximation that a homography applied to an image represents a viewpoint change inaccurate. We claim that in order to train detectors to work well in indoor environments, they must be robust to this type of geometry, and repeatable under true viewpoint change instead of homographies. Here we focus on the problem of detecting repeatable feature locations under true viewpoint change. To this end, we generate labeled 2D images from a photo-realistic 3D dataset. These images are used for training a neural network based feature detector. We further present an algorithm for automatically generating labels of repeatable 2D features, and present a fast, easy to use test algorithm for evaluating a detector in an 3D environment. 
\end{abstract}
\section{Introduction}
Feature detection and description have been an important and integral part of many computer vision applications related to inferencing 3D scene geometry, such as image retrieval, 3D reconstruction, and localization. In the literature, the performance evaluation of features have primarily been based on datasets that contain planar surfaces combined with affine transformations, regardless of whether the feature was constructed by hand (e.g., SIFT \cite{lowe2004distinctive, bay2006surf, rosten2008faster, shi1994good, harris1988combined} and their variants), or constructed by data-driven deep learning (e.g., \cite{LIFT}, \cite{detone2018superpoint}, \cite{savinov2017quad}, \cite{alcantarilla2012kaze}, \cite{DBLP:journals/corr/abs-1805-09662}). Although this is useful for making detectors and descriptors that work well on textured planar surfaces or arbitrary scenes with small viewpoint changes, modern applications of image features, like augmented reality, autonomous robot navigation, and building large-scale city maps, often further require features to work well with regard to large viewpoint changes within various geometries significantly different from planar surfaces.

When the existing features are evaluated in scenarios that contain significant viewpoint changes between non-planar surfaces, researchers have found that most methods perform poorly \cite{aanaes2012interesting,moreels2007evaluation}. One prominent exception is the MagicPoint detector \cite{detone2018superpoint}. DeTone \etal introduce a network architecture called SuperPoint and two networks, one named MagicPoint, which is a detector trained to recognize corner points with the use of a synthetic dataset. This dataset is comprised of simple geometric shapes, like grids and boxes, which was then augmented with noise. On this synthetic 3D dataset, it was able to achieve $0.979$ mean average precision, which is a  $>0.3$ mean average precision increase over traditional corner detectors like Shi \cite{shi1994good}, FAST\cite{rosten2008faster}, and Harris\cite{harris1988combined}. The other network is both a detector and a descriptor that they also call SuperPoint. The detector for SuperPoint, which we call SuperPoint-COCO to differentiate it from the architecture of the neural network, was trained using homographic adaptation, which is a process DeTone \etal used to adapt a set of detections to be viewpoint invariant on the MS COCO dataset. This is done by simulating viewpoint changes using homographies applied to the images. As the MS COCO dataset does not have ground-truth feature detections, the detection labels are provided by the MagicPoint detector. The SuperPoint-COCO detector is not pre-trained on the synthetic dataset and is only trained on the augmented MS COCO dataset. In our tests we show that SuperPoint-COCO suffers a large performance penalty compared to the MagicPoint detector in real 3D scenes. We believe this indicates that training on 3D data is necessary to develop viewpoint invariance with regards to non-planar surfaces. 

To this end, we developed methods to collect 2D images of 3D data and find the ground truth keypoints within these images to mimic the success of MagicPoint, but with data that more closely resembles scenes encountered in the use of these detectors. We generated a 3D data set using the Gibson simulator \cite{gibson} with the Matterport dataset \cite{Matterport3D} to tackle the issue of precision and variety in 3D data. We develop an algorithm for computing a repeatable set of detections given a set of 3D data and a collection of other detections, and we developed a fast, accurate, and powerful way to benchmark algorithm in indoor environments using the Scannet dataset \cite{dai2017scannet}. We will release the source code for all tools we developed.
\section{Previous Work}
Because the SuperPoint architecture used for MagicPoint was proven to be effective for learning viewpoint invariance on non-planar surfaces, we decided to use this network architecture, as well as the loss function, when selecting a network to learn our new dataset. The model is based on VGG \cite{vgg} and outputs a low spatial dimensional (1/8th scale) grid of values, storing the extra spatial data in the depth of the layer so it can be reshaped back to a full resolution interest map. A softmax is applied to each cell of the grid before it is reshaped to prevent double detections. The loss is the mean of a softmax cross entropy loss across each cell. We utilize the implementation by Rémi Pautrat and Paul-Edouard Sarlin \cite{superpointimpl}. 

Two papers evaluate the performance of detectors with regards to viewpoint invariance on non-planar surfaces.
Moreels and Perona 
\cite{moreels2007evaluation} conclude that the Difference of Gaussians detector (DoG) performs almost as well as the Hessian Affine detector. Aan{\ae}s \cite{aanaes2012interesting} utilizes the DTU Robot Dataset, to evaluate several detectors. Aan{\ae}s concludes that DoG, Hessian Blob, and Harris Corner detectors are also viewpoint invariant. However, neither of them test any data based detectors. 

% LIFT uses sparse 3D reconstruction, limiting their set of features to just what SIFT outputs
The most similar method to the one we develop is LIFT \cite{LIFT}. LIFT is trained to predict the subset of features detected by SIFT that are not flagged as outliers during 3D reconstruction. However, this can result in inaccurate labels. Due to imperfect feature matching, not every good point survives 3D reconstruction. Even worse, bad points can pass this check if not enough viewpoints capture it. Lastly the datasets are not very large as they use the Piccadilly and Roman Forum datasets \cite{wilson2014robust}. Our network utilizes a large dense 3D scene capture dataset. This allows us to initialize with a variety of features in addition to SIFT, and prevents inaccuracies in labeling as we can directly compute feature projections.

% LF-Net uses Scannet, which has limited viewpoints and they don't have our algo so they use 15 frame gaps
LF-Net\cite{DBLP:journals/corr/abs-1805-09662} trains using the Scannet Dataset. The Scannet dataset \cite{dai2017scannet} is comprised of videos captured by a handheld depth sensor within small indoor environments. Most of the sequences consist of several loops within a single room. LF-Net trains their network on 15 frame intervals with a single frame being projected into the next, which, by our calculations, yields an average angle change of $12.5^{\circ}$. The authors show that the Superpoint-COCO descriptors outperform both LF-Net and LIFT for indoor feature matching with wide baselines, but do not directly evaluate the reliability of detections.
%In their indoor matching test, their tests showed that SuperPoint outperformed all other feature detector and descriptors algorithms they compared, including LIFT \cite{LIFT}, A-KAZE\cite{alcantarilla2012kaze}, and

Rosten \etal used a small dataset comprised of 37 images with significant viewpoint changes to find the parameters for the FAST corner detector \cite{rosten2008faster}. Savinov \etal trains a network to output viewpoint invariant detection by training it to assign patches that correspond to the same point the same rank \cite{savinov2017quad}. They train on the DTU Robot Image Dataset \cite{aanaes2012interesting} as well as the NYUv2 dataset \cite{Silberman:ECCV12}, both of which are 3D indoor datasets. However, they only train on a handful of images (n=40) and test on fewer. In our work we train using a very large dataset of over 100,000 images from 150 different 3D building captures.

Many papers apply affine transformations to images to simulate viewpoint changes. There are many datasets \cite{hpatches, mikolajczyk2005comparison, cordes2013high, fischer2014descriptor} that are composed entirely of planar objects. There are also many datasets\cite{heinly2012comparative, zitnick2011edge, winder2009picking} that image objects far in the distance. The combination of far distance and small rotations make the images roughly planar and an affine transformation between images a reasonable approximation. This assumption was previously necessary due to the lack of availability of labeled 3D data. 
However, with the recent surge in 3D data coming from scene segmentation research \cite{Silberman:ECCV12, dai2017scannet, Matterport3D}, we can now train and evaluate detectors without this assumption on scenes where this assumption would not be appropriate.

\section{Proposed Technique}
\subsection{Training Set Generation} \label{sec:datasets}

%-------------------------------------------------------------------------
To generate a large amount of 3D data for training our network, we utilize the Gibson simulator, which renders photo realistic viewpoints of scenes captured with a Matterport sensor \cite{DBLP:journals/corr/abs-1709-06158}. The benefit of using simulated data is the ability to capture thousands of viewpoints from awkward angles, which is important to the method by which we generate ground truth labels for detection. Simulated data also has almost perfect pose and allows for more accurate calculation of ground truth for the training process. From the 340 areas, we select 150 areas that have a significant amount of texture and or objects filling the scenes because many areas are of completely empty and texture-less apartments. We capture images from each of the 150 areas by randomly positioning the camera within the bounds of the scene and checking if the camera position is valid. The orientation of the camera is randomly sampled with the yaw sampled uniformly from $[0, 2\pi)$, the pitch sampled from a Normal$(\mu=0,\sigma=\frac \pi 8)$, and the roll being sampled from a Normal$(\mu=0,\sigma=\frac \pi 4)$.
The validity of a camera position is determined by casting a ray in each axial direction and outward from the camera viewpoint and checking that it falls within acceptable bounds. The lower bound was set to be 0.6 meters for all rays to stop the camera from intersecting with walls, and the upper bound was set to be 5 meters for the top and bottom, 20 for the sides, and 10 meters for the viewpoint ray to keep viewpoints inside the model.

\begin{figure}[htb]
    \centering
    \includegraphics[width=0.7\linewidth]{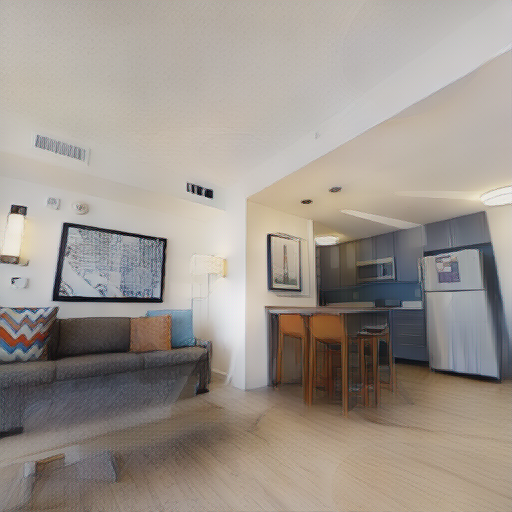}
    \captionof{figure}{%
    A room rendered within the Matterport dataset by the Gibson simulator \cite{gibson}. Note the slight amount of smearing on the table caused by Goggles filling in the data.
    }
    \label{fig:a}
\end{figure}

Part of the nature of the Gibson renderer is that it fills in the gaps for pieces of data that were missing, leaving blurry artifacts that shift around in some of the images, which can be seen in Figure \ref{fig:a}. Attempts to filter out these images using blur detection were made, but the nature of the images made it too difficult to discern good images from bad ones. However, the majority of the images are of good quality, and the model has shown the ability to transfer what it learned to the real world.

The bounding box around each area was used to estimate how many images should be taken for the area. However, we are more interested in the effective volume of the area, which we estimated by sampling 100 images to get an estimate for the percentage of image that are usable within the bounding box. We then multiply the volume given by the bounding box by the ratio of images we found to be usable. For each 1 meter cubed of effective volume in an area, we captured 10 images, for a total of 407389 images extracted from the 340 areas. 

\begin{figure*}[htb]
\includegraphics[width=\textwidth]{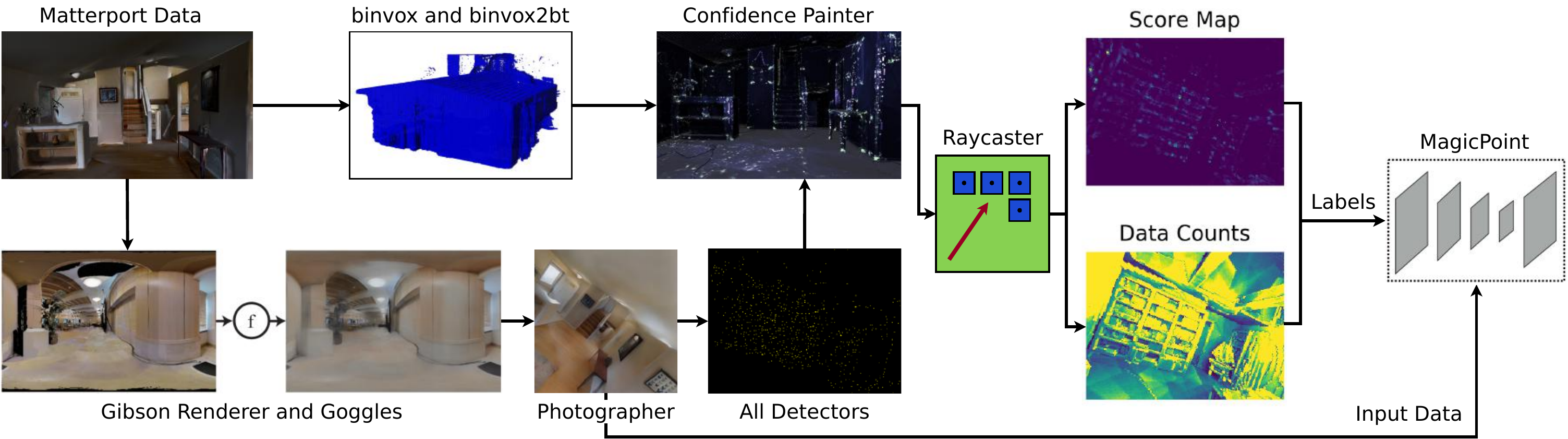}
\caption{Full overview of the method we used to train our network to detect points that are repeatable using 3D data. The process starts with using Matterport Data \cite{Matterport3D}, which is rendered using the Gibson renderer. The gaps are filled using a neural network called Goggles, and thousands of valid pictures are taken using the Random View Photographer. These images are passed to a collection of detectors to get a score for each pixel, which are then projected onto an octomap\cite{hornung13auro}, which is found by converting the mesh to an octomap using binvox\cite{binvox}. This painted map is then used to generate ground truth detections for the 2D image. These labels are then used to train SuperPoint\cite{detone2018superpoint}.}
\label{diagram}
\end{figure*}

\subsection{Ground Truth Labeling}
We propose a technique that, given a set of detections $D_i\subset \mathbb{Z}\times\mathbb{Z}\times \mathbb{R}_+$, on some set of images $I_i\in\mathbb{R}^{n\times n}$ with pose $P_i\in\text{SE}^3$, within an area that has a 3D model, can find the subset of detections within each frame that are viewpoint invariant. 
For any frame $I_i$, we can find the location of a detection $(I_x, I_y, c)\in D_i$ within the map by raycasting from $P_i$ in the direction $P_i\begin{bmatrix}
I_x & I_y & 1 & 1 
\end{bmatrix}^T$. We paint the first point hit in the map using the confidence of the detection $c$. We also track the number of images that have seen the pixel, which we call the number of views, for use in averaging later. Once this painting process is complete, we are left with a 3D map of all detections given, which can be seen Figure \ref{diagram}. 
Using this map, we can then evaluate whether the detection is repeatable within the set of detections by once again iterating through each frame $I_i$ and retrieving the values for confidence and number of views within the map at each detection$(I_x, I_y, c)\in D_i$. This is done by, once again, raycasting from the pose in direction of the pixel and retrieving the confidence value and number of views of the first point hit. We can then evaluate the repeatability of the detection based on how confident each view of the detection was about there being a feature at that point. In practice, there is quite a bit of noise involved in this process, so we have to perform additional preprocessing to reject noisy points and boost the signal of points that the original set of detections might not have detected.

\subsubsection{Map Creation Details}
Because our method is predicated on all repeatable detections existing in the set we are going to narrow down, we used all of the easily available detectors we had to generate the set of detections to help ensure that we had a comprehensive set of detections. We use SIFT\cite{lowe2004distinctive}, SURF\cite{bay2006surf}, ORB\cite{rublee2011orb}, Harris Corners, Good Features to Track (GFTT)\cite{shi1994good}, and MagicPoint\cite{detone2018superpoint}. We set the confidence of each detector to be $\frac 1 N$, where N is the number of features detected on that frame, to avoid over saturating the map with too many detections. 

To allow for fast raycasting within the map, we utilize an Octomap\cite{hornung13auro} and discretize the map, which was originally a mesh, to a voxel grid with a resolution 0.01 meters to aggregate the 3D detections in space \cite{binvox}\cite{nooruddin03}.
Calculation of the score maps is an expensive calculation that takes approximately 500
 core hours. However, it is easily parallelizable, so we utilized the Pacific Research Platform cluster computer to calculate the ground truth, allowing for the calculation to be performed in 6 hours. 

\subsubsection{Extracting and Filtering of Points}\label{technique:eval}
The problem with evaluating each of the detections is the amount of noise that can be present in the map. 
The most important source of noise is the noise caused by a phenomenon we call backshadow, which is caused by noise in the projection of 2D data into 3D space can cause a shadow like effect on regions far away from the region in 3D space the projection was meant to effect. If each point in 3D space is evenly viewed, then backshadow causes few issues because each point would have enough data to reject outliers, like those caused by backshadow. This is why the ability for simulated data to capture odd viewpoints is important. However, even with our simulation, bias caused by the position checker caused there to occasionally be areas with few views, and backshadow is a serious issue for points with a low number of views. As a result, when compiling the score map, we reject the use of the score map and use MagicPoint as the ground truth if the number of views is lower than the threshold=10 at any individual point. 
For the rest of the points, we want to utilize the map of the number of views that viewed each pixel, which we call counting map, to divide the original score map we obtain so we can find the average value of the score map for each pixel. However, the problem with the counting map is that it can have spiky regions, which can cause abnormal averages. To counter act this, we erode the counting map and blur it with kernels of size 9 to smooth the counting map, which we then use to divide the score map to obtain the mean score map.

Once we have obtained our mean score map, we need to find the set of repeatable detections. However, using the mean score map to evaluate the original set of detections yield numerous double detections for the same point. The first problem was that detections would often yield slightly different locations for the same detection. This led into the second problem, which was that the discretization of the map would cause all of these points to fall under the same voxel and all be assigned a high score within the mean score map. In theory, we would want the detections for a single 3D feature to lie within the same space within the discretization of the map. However, we also need a high enough resolution to distinguish between close points. 

\subsubsection{Filtering Double Points}
We propose the following solution to this problem. We create a set of candidate points, which we filter to remove double points, from which we select a subset using the mean score map.
First, we wanted to to add points to the original detection set using the map because we found it to be far from perfect. To obtain candidate points from the mean score map, we applied Difference of Gaussians (DoG) to the score map and obtained the peaks greater than 0.01. However, the discretization of the map at close distances would present many detections, so we prevented the map from adding to the set of detections if the detection was closer than 0.5m, which we checked by utilizing a depth map for the frame.

We then compiled all of the detections into a single map of detections. Let $D_s$ be the set of detections from the mean score map and $D_a$ be the set of detections from all other detectors. Let $S$ be the mean score map. Then the map of candidate detections $C$ was set to 
\begin{equation}
C_{x,y} = \begin{cases}
3 & \text{if } (x, y) \in D_s\cap D_a \\
2 & \text{if } (x, y) \in D_s\setminus D_a \\
1 & \text{if } (x, y) \in D_a\setminus D_s \\
0 & \text{else }
\end{cases}
\end{equation}
Now that we had a map of the priority of each detection, we could filter the double points by applying a Gaussian blur to $C*S$ and extracting the local maxima (with a radius of 2px). We multiply the priority map by the score map to bias the candidate detections to be in locations with a high score because we found that the map would often present a more accurate location for the feature than any of the feature detectors had found. These peaks are then assigned values from the score map to allow for accurate thresholding, which we set to 0.05, creating our final set of detections.

\section{Experiments}
\subsection{Test Set Selection}
We did not want to use the Gibson dataset to test algorithms because of issues caused by Goggles\cite{gibson}, the algorithm used to fill holes in the dataset, as it can distort the images. However, due to our desire to test the ability for a network to handle non planar perspective changes, we needed a corresponding testing dataset that has depth, otherwise the testing method will not be able to check where a feature point lies in 3D space. 
There are many datasets and tests that exist in the detection literature. However, as mentioned in the previous works section, most viewpoint invariance tests are based on datasets that approximate affine transformations on planar surfaces. There two tests designed to check viewpoint invariance in non-planar surfaces though: the test by Aan{\ae}s that utilizes the DTU Robot Dataset\cite{aanaes2012interesting}, and the test by Moreels and Perona \cite{moreels2007evaluation}.Unfortunately, Moreels and Perona \cite{moreels2007evaluation} do not make their evaluation method available and the test developed by Aan{\ae}s \etal\cite{aanaes2012interesting} takes multiple days to weeks to evaluate a single detector.

As a result, we designed out our detector repeatability test, which is based on the Scannet\cite{dai2017scannet} dataset. The Scannet dataset is comprised of videos of indoor environments captured by a mobile scanning device. The dataset consists of 707 small indoor areas, like a doom room, each of which can feature multiple videos, for a total of 1513 RGB-D videos of indoor environments. The original purpose of the videos was to reconstruct the rooms, so each of the scans typically capture a person walking around in a circle, moving the sensor outwards to scan the room, then moving to scan other objects within the room, like sofas. 

The Scannet dataset does not fulfill our requirements for the training set because we need many different perspectives of the same objects for our algorithm to work properly. However, as a test set, it is quite good. The motion of the camera typically shows objects at angles between 0\degree and 90\degree because of the scanning motion, which we consider to be sufficient. Angles further than 90 degrees start to introduce issues with occlusion, where the object itself might occlude the visibility of it's corners, which requires the kind of very precise depth that is hard to find in any modern dataset.

The full Scannet Dataset is very large, so we restrict our usage to the first 200 areas and only use every 30th frame. We selected this framerate from a visual inspection of the data, which showed that every 30th frame represented an average of 12.5 degrees of viewpoint change. The framerate did not seem to impact our results that much because we perform and all images to each other in an area, which limits the impact of such a choice. To perform this pseudo all-to-all comparison, we first cache a map of the pairwise correspondence between the frames that describe whether images view the same area of the scene. We check whether a query and a candidate image capture the same area by using the depth from the candidate image to project points into 3D space, which we then project back into the query image to check how much of the image contains points from candidate. If at least 10\% of the candidate is contains the backprojection of the query, the pair of images is considered to be worth processing. This results in an average of 40-90 degrees of viewpoint change, depending on the area. 
\subsection{Detection Repeatability Test}
To measure a feature detector's repeatability with respect to viewpoint change, we want to know how often it is able to predict the same point from multiple viewpoints. 
However, the feature detector is probably not going to be able to detect exactly the same point because it isn't perfect. Therefore, we want to know if the feature detector is able to detect the same point within some distance. However, choosing a distance threshold is an easy way to add a highly influential variable to testing that can obscure the actual quality of an algorithm. Instead, we compile a histogram of the distances to the closest point for each detection. The histogram has the advantage of being both easy to interpret while maintaining the transparency in how many points a detector is detecting. A degenerate detector, for example, is perfectly repeatable but would show up as a huge spike at 0 distance. 
The problem is that some detectors, like the Hessian Affine detector, detect around 1000 features, which is an order of magnitude more points than detectors like Good Features to Track or MagicPoint \cite{aanaes2012interesting}. Although we could just allow these detectors to detect their max number of points, it would cause scaling issues with the histogram and limit the interpretability of the data. Therefore, we limit the number of detections to 2000 and apply non maximal suppression to prevent detectors from predicting double points. Some methods only  plot a log histogram of the number of detections for each distance. This allows us to fairly test other algorithms while keeping the histogram manageable. To deal with the scaling of the histogram, and to make the data more interpretable, we also include a plot of the percentage of detections for each distance.

To calculate this histogram, we take the detections from some image of size ($240\times 320$), which we call the candidate, and back project it into the query image. This gives us the position of the detections found in the candidate in the query image. We can then find the distance to the closest candidate detection for each detection in the query image. To back project the detections from the candidate to the query image, we take each detection $a = (a_x,a_y)\in D_c$ and project it into 3D space using the candidate depth map $d_c\in\mathbb{R}^{n\times m}$. We can then use the shared camera matrix $K$ and the relative rotation $R$ and translation $T$ between the two cameras to project this detection back into the detection into the query viewpoint to get the candidate detections $D_c^q$. We then check whether the detection is obscured in the query frame as well as filter out noise in the depth by checking that the distance to the detection is close to the depth of the query image $d_q\in\mathbb{R}^{n\times m}$. Formally,
\begin{align}
\text{Let } &b_z\begin{bmatrix} b_x & b_y & 1\end{bmatrix} = 
K \begin{bmatrix}
R & T\end{bmatrix}
\begin{bmatrix}
a_x \\
a_y \\
d_{c}(xy) \\
1 \\
\end{bmatrix} \\
	D_c^q &= \{(b_x, b_y) \text{ if } |b_z - d_{q}(b_x, b_y)| < \epsilon\} \\
    H_{qc}(r) &= \sum_{c\in D_q} \begin{cases}
    1 & \text{if } |\min\{\|b-c\| ,b \in D^q_c\} - r|<0.5\\
    0 & \text{else} \\
    \end{cases}
\end{align}
Where $H_{qc} :\mathbb{N} \to \mathbb{N}$ is the histogram that maps the integer pixel distance from the query detection to the number of queries with that distance. Unmentioned in this formal description is the fact that we store the number of the unmatched detections in the last column of the histogram. For these unmatched detections, we only include the detection repeatability for points that are visible in both frames. 

One of the concerns with using 3D data in feature evaluation is that it has been exceptionally slow, with the DTU Robot Dataset test taking 10 seconds per a pair of images. If we restrict the dataset to just be comparisons within the same lighting conditions, that would be 424800 pairs, which would take approximately 50 days to evaluate. 
The datasets can also be huge, with the DTU Dataset taking 500 GB of storage. With this in mind, we developed a C++ code base that can evaluate a detector in 10 minutes on an Intel i9-7900X with 20 threads. We find the nearest neighbor of each query within the candidate by convolving a binary picture of the candidate detections $C$ within the query with a series of circular distance filters. Each of these filters $f_i:\mathbb{R}^{m\times n}\to \mathbb{R}^{m\times n}, i\in\{1, \ldots 20\}$ has a circle with some radius $i$, and we set a distance map $D\in \mathbb{R}^{m\times n}$ within the query frame to be equal to the minimum value of the $i$ such that $f_i$ is not zero at that point. Formally,
\begin{align}
f_i(x, y) = \begin{cases}
1 & \text{if } |\|(x, y)\| - i| < 0.5 \\
0 & \text{else}
\end{cases} \\
D(x, y) = \min\{i | f_i(x, y)\circ C \geq 1\}
\end{align}

\begin{figure*}[ht]
\centering
\includegraphics[width=\textwidth]{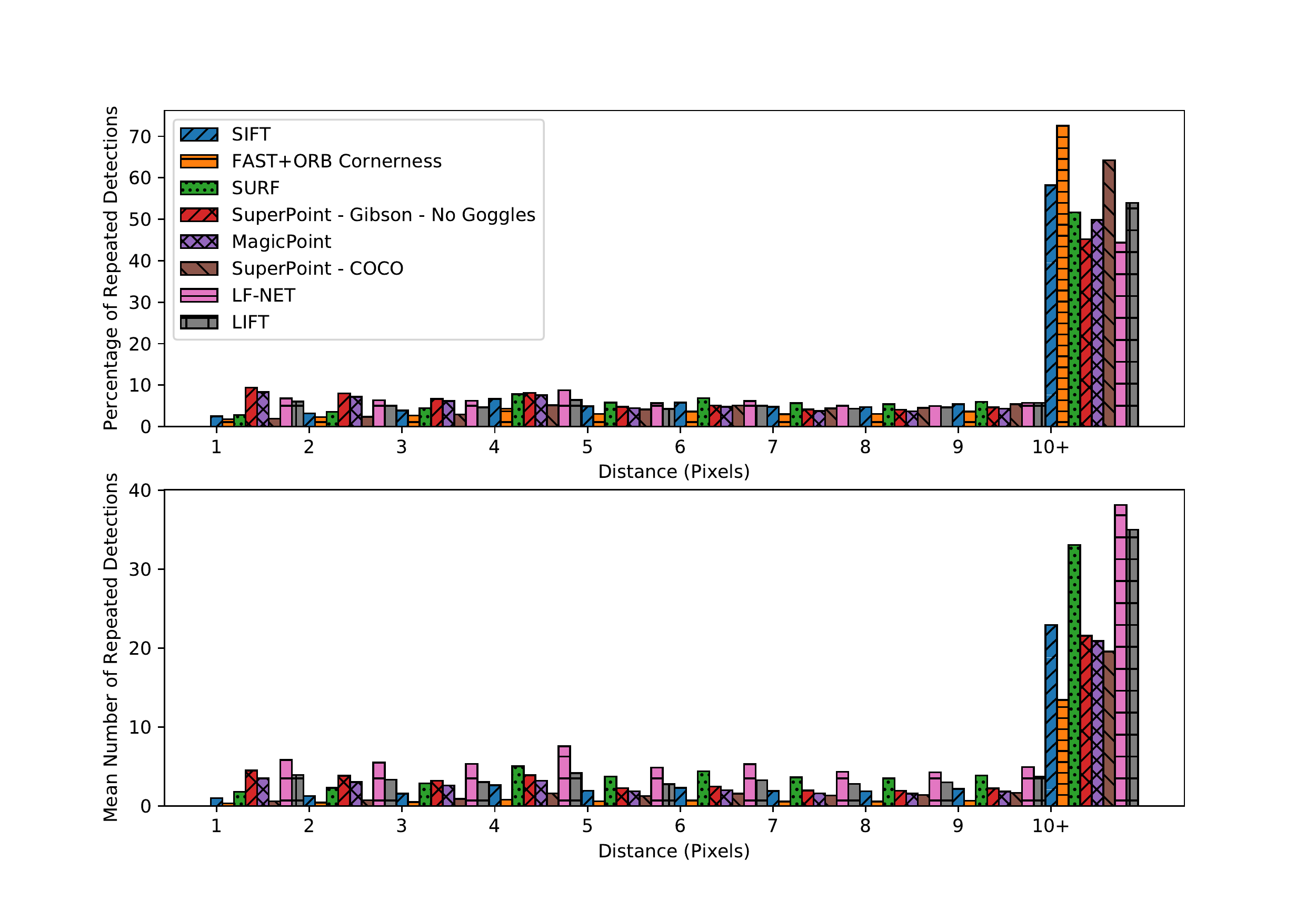}
\caption{Histograms of the average percentage (top) and number (bottom) of points whose closest point in the paired image is within a given distance range. This average is computed across all pairs in our test set. Higher numbers at lower distances means that the detector is more precise when repeating detections. The distance bin of 10+ indicates the percentage of detections that were not repeated.}
\label{results:hist}
\end{figure*}
\subsection{Results}
We compare the FAST\cite{rosten2008faster} detector with the default parameters from ORB\cite{rublee2011orb}, SURF\cite{bay2006surf}, SIFT\cite{lowe2004distinctive}, MagicPoint, SuperPoint,\cite{detone2018superpoint}, as well as our algorithm, which we label Superpoint-Gibson. 
The results of this comparison are shown in Figure \ref{results:hist}. The repeatability of a detector can be expressed as the percentage of points that have a corresponding detection within a certain distance. However we also want our detector to reliably detect as many points as possible. To this end, we create histograms of both the average number and percentage of points that have a corresponding point at a given distance. The 10+ distance bin can be considered points without a repeated detection. 

MagicPoint and our network, Superpoint-Gibson, outperform the other algorithms, detecting both more points and a higher percentage of points at a distance of 1,2, and 3 pixels than any of the other detectors. In addition, they both also predict a smaller percentage of unmatched points than the other detectors. Note that both MagicPoint and our algorithm were both trained on synthetic 3D data. While MagicPoint had a precisely labeled ground truth thanks to using a variety of simple shapes for which the labels were known, our algorithm was able to surpass MagicPoint by utilizing richer 3D data to create a new set of repeatable points. 

Similar to how SuperPoint-COCO was trained, our method incorporates detections from the MagicPoint detector as part of it's set of detection labels, but does not train on the synthetic dataset. The results from SuperPoint-COCO show that it is not enough to simply utilize the detections from a viewpoint invariant detector to inherit the viewpoint invariance, however our algorithm avoids the drop in performance by training using real viewpoint changes.

If we sum the bins from 0-3 pixels we get the number of points with a corresponding feature detection within 3 pixels. Our algorithm detects an average of 10.75 repeatable features per image. This represents a $13\%$ improvement over Magipoint which detects an avarage of 9.52 repeatable features per image. The performance of MagicPoint over SuperPoint-COCO hints at the relative importance on training on real viewpoint change to training on real images. By incorporating real images as well as real viewpoint changes we were able to further improve performance over SuperPoint-COCO, specifically when considering those features that can be detected reliably within 3 pixels. 

\section{Conclusion and Future Work}
Our main finding with respect to generating a synthetic dataset is that trying to fill in regions of images for which there is no data is detrimental to the training of feature detectors. The use of a neural network, such as Goggles\cite{gibson}, causes too many errors to be used in such a precise application, even to the point of occasional objects going missing from scenes. In future work on generation of synthetic datasets, we think that it would be best to mask out sections of terrain for which there is not data instead of filling it in from the void. With such a system in place, it might be possible to utilize lower quality datasets to allow for the diversification of data to include more cluttered scenes, outdoor scenes, lighting changes, and more.

The testing algorithm we designed is transparent and easy to use. Although the results are not as simple as assigning a number to a detector to rank how "good" it is, we believe that the design of the output allows for a less biased depiction for how well each detection performs so it's strengths can be seen. We once again confirmed the findings of previous benchmarks and showed that most detectors perform poorly with regards to viewpoint change relative to non-planar surfaces \cite{aanaes2012interesting,moreels2007evaluation}.

We showed that training on 3D data is a solution to designing detectors that are non-planar viewpoint invariant. Not only did we demonstrate that the same model when trained with 3D data outperforms the same model trained to be invariant to homographies applied to images, we also demonstrated a method of extending this training method to more environments. With a greater diversity of data, our algorithm for ground truth generation has the potential to train all purpose detectors that would be able to achieve very high precision detections, allowing for precise rigid body transformation estimation and more. 
{\small
\bibliographystyle{ieee_fullname}
\bibliography{egpaper.bib}
}

\onecolumn
\appendix
\section{Example Detections}
\begin{figure*}[ht]
\centering
\includegraphics[width=0.9\textwidth]{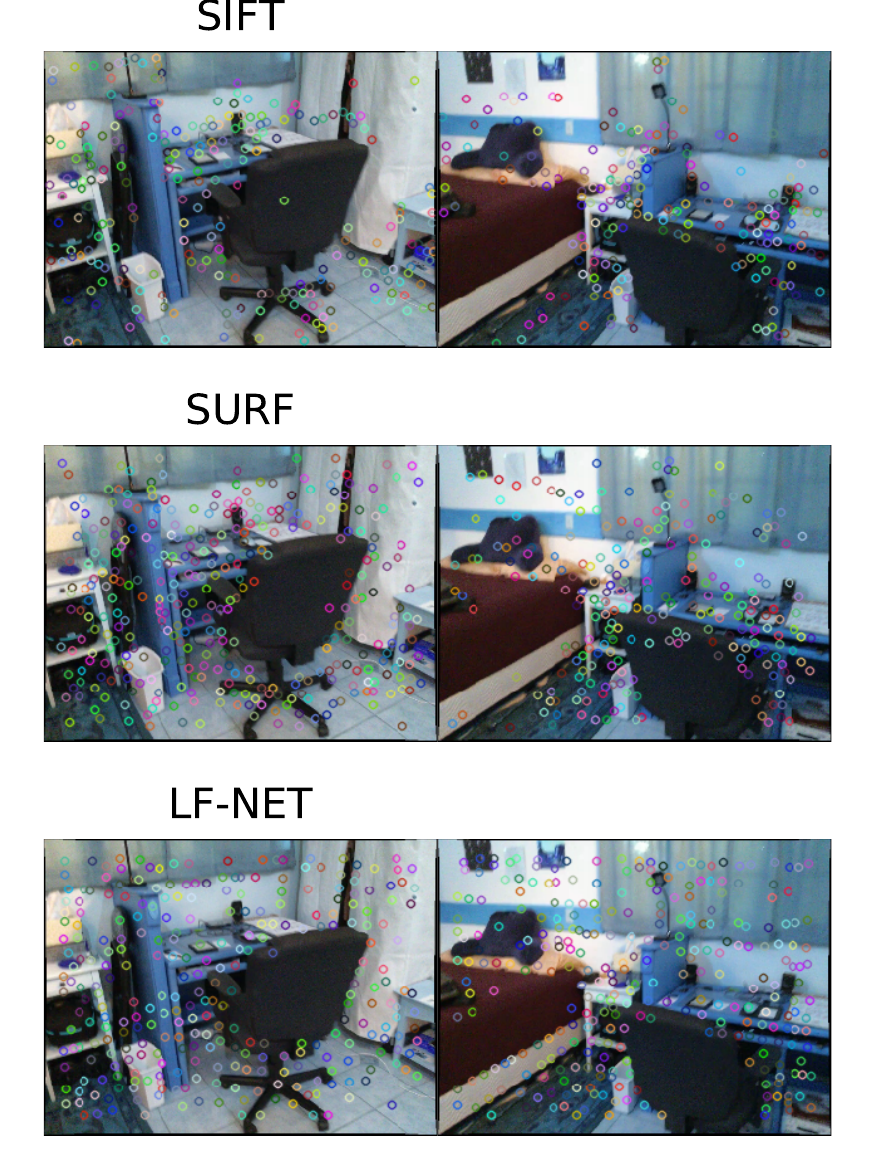}
\end{figure*}
\begin{figure*}[ht]
\centering
\includegraphics[width=0.9\textwidth]{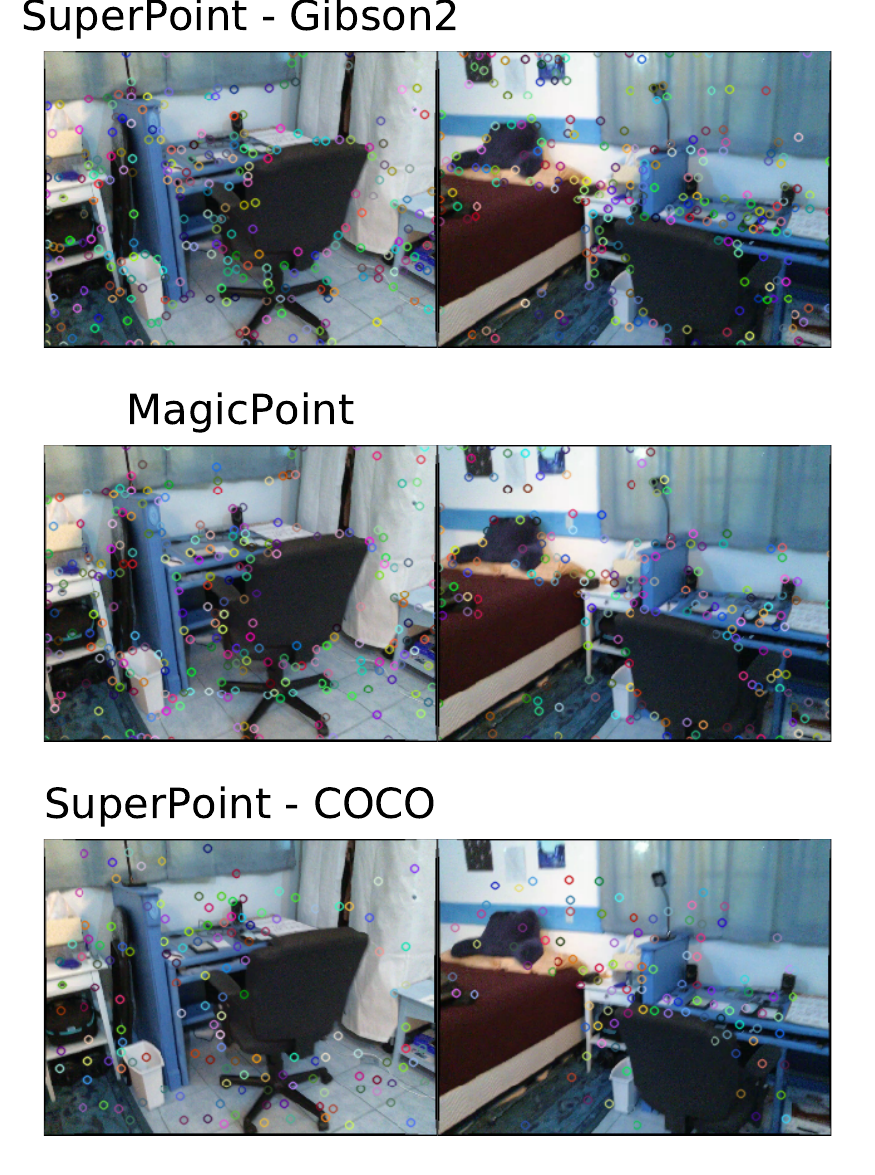}
\caption{Note how all corners of devices on the table are detected. Note also that our method detects the corners on the printer under the cabinet as well as the monitor stand. However, other methods fail to detect any of these points.}
\end{figure*}
\end{document}